\documentclass[9pt, conference]{IEEEtran}
\usepackage{makecell}
\usepackage{enumitem}
\usepackage{xspace}
\usepackage{graphicx}
\newcommand{\agent}[1]{\textit{#1}\xspace}

\usepackage{soul}
\usepackage{booktabs}
\usepackage{dblfloatfix} 
\usepackage{cuted}    
\usepackage{capt-of}  

\usepackage[T1]{fontenc}
\usepackage{xcolor}
\usepackage{pifont}   
\usepackage{wasysym}  

\newcommand{\good}{\textcolor{green!60!black}{\ding{51}}}    
\newcommand{\medium}{\textcolor{orange!80!black}{\LEFTcircle}} 
\newcommand{\bad}{\textcolor{red!70!black}{\ding{55}}}       

\definecolor{codebg}{HTML}{F7F7F7}

\AtBeginDocument{%
  }

\usepackage{url} 

\begin{document}

\title{AgRefactor: Self-Evolving Agentic Workflow for HLS Compatibility and Performance}

\IEEEoverridecommandlockouts
\author{
\IEEEauthorblockN{Yang Zou\textsuperscript{1,*},
Zijian Ding\textsuperscript{2,*},
Yizhou Sun\textsuperscript{2},
Jason Cong\textsuperscript{2}\thanks{\textsuperscript{*}These authors contributed equally.}}
\IEEEauthorblockA{\textsuperscript{1}Carnegie Mellon University \qquad
\textsuperscript{2}University of California, Los Angeles \\
yangzou@andrew.cmu.edu, \{bradyd, yzsun, cong\}@cs.ucla.edu}
}



\maketitle

\begin{abstract}
High-Level Synthesis (HLS) provides a fast path from concepts to silicon, but converting real-world software into synthesizable HLS code remains challenging due to restrictive language support and the gap between software and hardware programming practices. Existing automated and LLM-based refactoring approaches partially address this problem, yet they often lack flexibility, struggle to scale, and incur high computational costs. We introduce \textsc{AgRefactor}, an LLM-based multi-agent workflow for refactoring software into HLS-compatible programs. \textsc{AgRefactor} incorporates a self-evolving memory system that accumulates and retrieves factual and strategic knowledge across tasks, improving robustness and efficiency on unseen programs. To reduce cost and enhance scalability, it integrates automated refactoring tools, enabling agents to balance LLM-driven rewrites with efficient tool-based transformations. On 9 out of 11 challenging real-world benchmarks, which are $5-10\times$ longer than the most complex cases studied in prior work, \textsc{AgRefactor} outperforms or matches the state-of-the-art automated refactoring tool and a strong LLM-based baseline built on the same framework backbone. Further agentic performance optimization yields a 6.51$\times$ geometric mean speedup over the SoTA pragma tuning tool and a 1.20$\times$ speedup over optimized open-source designs with less than 20\% extra resources. \textsc{AgRefactor} is fully-automated and open-sourced.
\end{abstract}

\section{Introduction}

High-Level Synthesis (HLS)~\cite{cong2011high,cong2022fpga,scalehls,streamhls,ye2024hida,sisyphus,allo} was introduced to streamline the process of turning design concepts into silicon implementations, by raising the abstraction level from register-transfer level (RTL) design to C/C++-based HLS designs. Practical HLS design flows begin with a tedious step: refactoring the software repository into an HLS-compatible program~\cite{heterorf,heterogen,collini2024c2hlsc}. As reported in a real-world acceleration project of high energy physics~\cite{wojenski2024hardware}, this preparatory refactoring phase can take domain experts \emph{days} to produce an initial HLS-compatible program for further optimization. This highlights the need for more automation to reduce this labor-intensive task.

To address these difficulties, automated refactoring methodologies~\cite{heterorf,heterogen} have shown promise in converting software programs into HLS-synthesizable code. These approaches typically combine software engineering techniques such as dynamic invariant analysis and rely on fixed templates to rewrite non-synthesizable constructs. The increasing capabilities of large language models (LLMs) in software engineering motivate recent studies on using LLMs for HLS refactoring~\cite{hlsrewriter,collini2024c2hlsc}. These methods typically feature an agentic workflow that identifies non-synthesizable constructs and applies iterative refactoring based on compiler feedback. HLSRewriter~\cite{hlsrewriter} introduces retrieval-augmented generation (RAG) to store factual knowledge and rewrite templates, while C2HLSC~\cite{collini2024c2hlsc} first extracts each function and performs kernel-level rewriting with automatic testbench generation.

\begin{table}[h]
  \centering
  \caption{\small \textsc{AgRefactor} achieves better refactoring performance while reducing cost compared with current state-of-the-arts.}
  \vspace{-5pt}
  \label{tab:baseline-comparison}
  \begin{tabular}{lccc}
    \toprule
    \textbf{\small Method} & \textbf{\small Generalizability} & \textbf{\small Scalability} & \textbf{\small Cost}  \\
    \midrule
    Rule-based~\cite{heterorf} & \bad\ Poor & \good\ Good  & \good\ Good  \\
    LLM-based~\cite{hlsrewriter} & \medium\ Medium & \medium\ Medium & \bad\ Poor   \\
    \textsc{AgRefactor} & \good\ Good & \good\ Good & \medium\ Medium \\
    \bottomrule
  \end{tabular}
  \vspace{-10pt}
\end{table}

Existing methods face three key challenges in HLS refactoring: generalizability, scalability, and cost. To mitigate these challenges, we propose \textsc{AgRefactor}, an LLM-based multi-agent workflow capable of refactoring real-world software programs into HLS-compatible programs. To address the issue of generalizability, we develop a self-envolving workflow with a specialized agentic memory system. Factual and strategic knowledge is automatically extracted and retrieved as the agents interact with diverse tasks (programs) and HLS tools. As the system accumulates knowledge from past experience, \textsc{AgRefactor} demonstrates higher efficiency and robustness when applied to unseen software programs. Unlike manually crafted RAG approaches, our memory system adapts to the model's specific failure and success modes, providing targeted knowledge retrieval that improves refactoring accuracy and efficiency. Relative to general agentic memory systems~\cite{lewis2020rag,packer2023memgpt,gutierrez2024hipporag,wang2024agentworkflow,wang2024memoryllm,xu2025amem}, we create a specialized knowledge format, dedicated interaction patterns, and a novel distance metric, all designed to achieve superior performance on this task.

To alleviate the challenges of scalability and cost, we leverage the tool-calling capability of LLM agents. Specifically, we equip \textsc{AgRefactor} with HeteroRefactor~\cite{heterorf} (abbreviated as HeteroRF in figures and tables), the state-of-the-art (SoTA) automated open-source refactoring tool for HLS. Naturally, for programs that can be fully handled by tools, we avoid the computational overhead of invoking LLMs. Furthermore, our framework allows LLM agents to balance between code generation and tool calls: they focus on removing or rewriting software constructs unsupported by HeteroRefactor, while delegating the remaining work to the tool. As a result, our tool-enhanced agentic workflow combines the flexibility of LLM-based code rewriting with the scalability and low cost of automated refactoring.

While recent search-based approaches like AlphaEvolve~\cite{alphaevolve} utilize scalar feedback (e.g., latency) to drive code mutations, \textsc{AgRefactor} provides the agent with richer information from raw EDA logs and scheduling reports. This capability allows the agent to pinpoint specific critical-path bottlenecks. Furthermore, we integrate a multi-tool environment that enables the agent to bootstrap from established human knowledge rather than performing blind mutations.

Our contributions are summarized as follows:
\begin{enumerate}
\item We design and implement \textsc{AgRefactor}, an agentic workflow capable of refactoring real-world software into synthesizable, high-performance HLS code. Unlike existing approaches that focus solely on correctness and pragma insertion, we distinctly address structural refactoring, which most significantly impacts hardware performance.
\item We propose a self-evolving agent memory system specialized for this task. By continuously accumulating and retrieving both factual and strategic knowledge across varying tasks, our workflow enhances the base LLM's robustness in HLS refactoring while requiring minimal human intervention.
\item We integrate automated refactoring tools into our agentic framework, creating a hybrid pipeline powered by both LLM agents and an algorithmic backend. This integration significantly reduces computational costs and improves the scalability of LLM-driven HLS refactoring.
\end{enumerate}

Evaluation on diverse real-world cases shows \textsc{AgRefactor} outperforms or matches both the state-of-the-art automated refactoring tool and a strong LLM-based baseline on 9 out of 11 benchmarks. Our optimization agent achieves a 6.51$\times$ geometric mean speedup over SoTA pragma tuning tools AutoDSE~\cite{autodse} and a 1.2$\times$ speedup over highly optimized open-source HLS designs with less than $20\%$ extra resource.

\section{Motivation}
\label{sec:motivation}

\begin{figure}[t]
    \centering
    \includegraphics[width=0.85\columnwidth]{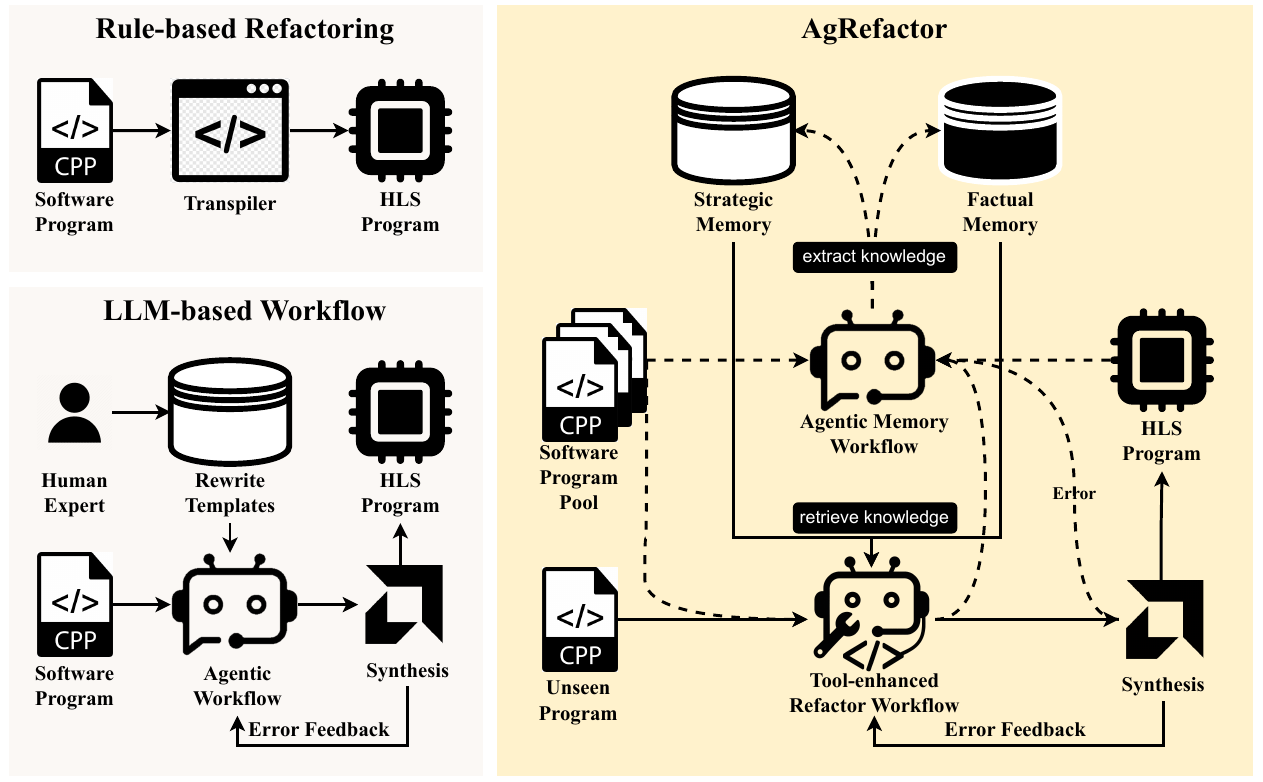}
    \caption{\small Limitations of existing methods motivate our workflow design. (1) Automated refactoring suffer from poor generalizability. State-of-the-art automated refactoring cannot handle external libraries and complex struct data types. (2) LLM-based workflows require manual effort to build a RAG system and are subject to the intrinsic randomness of LLMs, leading to high variance in outputs. Moreover, pure LLM-based refactoring incurs significant cost. In contrast, \textsc{AgRefactor} addresses these challenges through a tool-enhanced, self-evolving workflow, enabled by an agentic memory system (Sec.~\ref{sec:overview}).}
    \label{fig:motivation}
\end{figure}

\begin{table*}[t]
  \centering
\caption{A comprehensive benchmark suite with complex real-world examples. Our benchmarks cover real-world programs from image/video codecs, cryptography, and computational genomics.}
\vspace{-5pt}
  \small
  \begin{tabular}{lll}
    \toprule
    \textbf{Category} & \textbf{Name} & \textbf{Lines of code}\\
    \midrule
    leetcode~\cite{leetcode}       & \makecell[l]{
        maxSlidingWindow, delNodes, wordBreak, getSkyline, canDistribute criticalConnections, \\
        smallestRange, solveSudoku, removeInvalidParentheses, swimInWater
    }
                   & 21-67\\ \midrule
    HLSRewriter~\cite{hlsrewriter}    & \makecell[l]{
        regUpdate, reverseKGroup, trap, FibSubseq, knn, rotateGrid, maxPoints, long,\\
        qrd\_compute, forward, dfs, aes\_encrypt
    }
                    & 10-115\\ \midrule
    C2HLSC~\cite{collini2024c2hlsc}   & \makecell[l]{
        Frequency, CumulativeSums, Overlapping, quicksort, \\
        present80\_encryptBlock, sha256\_update, des\_crypt
    } & 11-221\\ \midrule
    HeteroRF~\cite{heterorf} & \makecell[l]{
        dfs, mergesort, linkedlist, ahocorasick, strassen
    } 
                   & 65-304\\ \midrule
    libsodium~\cite{libsodium}      & \makecell[l]{
        blake2b\_compress, chacha20\_stream, argon2\_fill\_segment
    } & 126-370 \\ \midrule
    minimap2~\cite{minimap2_repo}       & \makecell[l]{mm\_chain\_dp\_orig} & 473 \\ \midrule
    libjpeg-turbo~\cite{libjpeg_turbo}  & \makecell[l]{
        encode\_one\_block, median\_cut, select\_colors, fill\_inverse\_cmap, idct\_generic
    }
                   & 279-\textbf{754} \\ \midrule
    av1~\cite{aom_av1}        & \makecell[l]{
        av1\_apply\_temporal\_filter, av1\_compound\_type\_rd
    } & 407-\textbf{1266} \\ 
    \midrule
  \end{tabular}
  \label{tab:benchmark}
\end{table*}

In this section, we motivate our workflow design by evaluating existing approaches with a comprehensive benchmark and highlighting their limitations. Existing approaches are usually evaluated on a small number of manually constructed test cases --- typically ones that can be readily solved by their framework. To more thoroughly examine the limitations of state-of-the-art approaches, we evaluate them on an extended set of real-world benchmarks, which includes multiple standalone files with up to 1000 lines of code. The details of our benchmark suite are provided in Table~\ref{tab:benchmark} and Section~\ref{sec:benchmark}. Since existing workflows are not fully open-sourced and are tailored specifically to Catapult HLS tools, we make our best effort to reproduce their approach.

\begin{figure}
    \centering
    \includegraphics[width=0.85\linewidth]{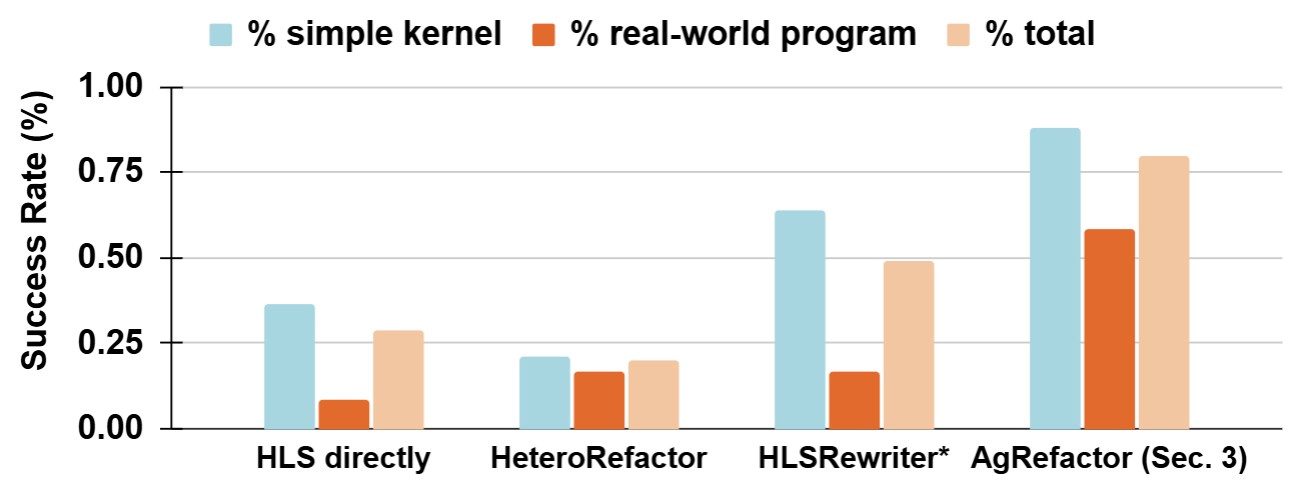}
    \caption{\small Summary of one-shot refactoring success rate across different workflows. We adapt the HLSRewriter methodology onto our framework as a controlled ablation (see Section \ref{sec:eval_setup} details). Each baseline is tested with 45 kernels. The category ``real-world program'' includes amalgamated files from real-world software repositories together with the ``strassen'' kernel, while ``simple-kernel'' covers the remaining benchmarks listed in Table~\ref{tab:benchmark}. All LLM-based workflows use ``GPT-5-mini''. Existing methods fall short in refactoring real-world benchmarks.}
    \label{fig:motivation_result}
\end{figure}

Figure~\ref{fig:motivation_result} summarizes the results. Although both HeteroRefactor and HLSRewriter perform well on their own benchmarks, they failed many more larger benchmarks, such as libjpeg-turbo. Beyond the known limitation of handling external libraries such as STL containers (e.g., ``std::set'' and ``std::map''), HeteroRefactor also shows limitations when dealing with common pointer operations. In particular, it cannot process nested pointer arithmetic ($A\rightarrow B\rightarrow C$) correctly.

State-of-the-art LLM workflows, even those backed by GPT-5, show limited success on real-world repositories (Fig. \ref{fig:motivation_result}). Three main challenges emerge: (1) real-world programs contain incompatible implementations missing from standard knowledge bases, (2) refactoring fails when invalid patterns span multiple locations in long programs, and (3) iterative error feedback makes LLMs slow and computationally expensive compared to automated tools. Furthermore, our analysis reveals LLMs frequently misuse unsupported C++ features (e.g., lambda functions) or incorrectly expose internal structs to hardware interfaces.

Because LLMs make random mistakes, anticipating them via manual prompt engineering requires impractical effort. Instead, robust refactoring requires a task-specific plan that properly decomposes the process. To this end, we propose a self-evolving agentic workflow that automatically accumulates factual and strategic knowledge from diverse tasks, retrieving it to guide agents on new programs.

Since the original HLSRewriter targets the Catapult HLS toolchain and its key components (the RAG knowledge base and identifier prompts) are not open-sourced, a faithful reproduction is infeasible. We therefore construct a strong baseline that adapts the HLSRewriter methodology onto our own framework, as detailed in Section~\ref{sec:eval_setup}.

\section{Methodology}

\subsection{Framework}
\label{sec:overview}

\begin{figure*}
    \centering
    \includegraphics[width=0.85\textwidth]{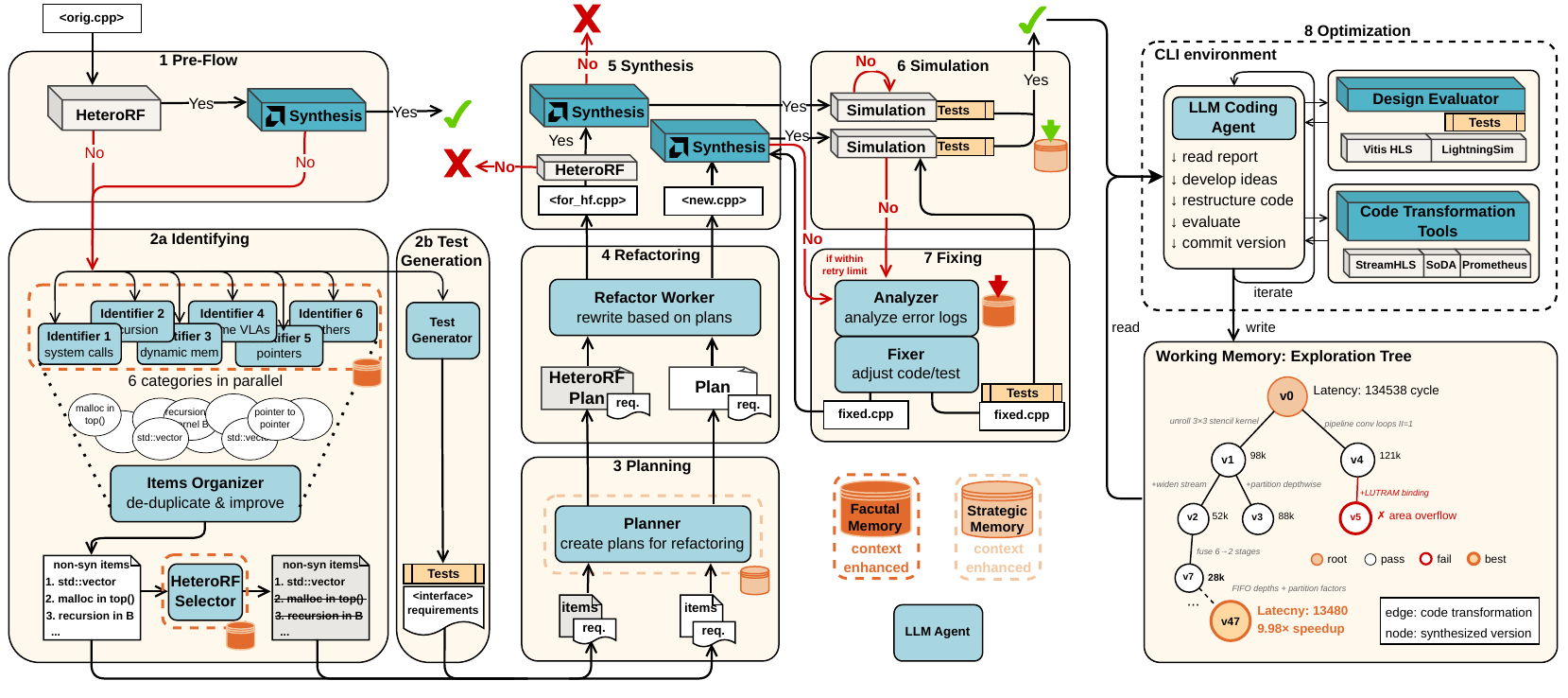}
    \vspace{-10pt}
    \caption{Overview of \textsc{AgRefactor}. Given a C/C++ program and a user-specified top-level function, the framework automatically produces a synthesizable HLS implementation by progressing through identifying, planning, refactoring, and fixing stages, while continuously updating a long-term memory bank. Once refactored, the synthesizable code and its testbench are forwarded to a performance optimization agent. Situated in a CLI environment with access to evaluation and code transformation tools, this agent leverages an internal tree-based working memory to systematically explore, apply, and track diverse code optimization strategies for optimal performance.}
    \vspace{-10pt}
    \label{fig:overview}
\end{figure*}

Figure~\ref{fig:overview} shows the overall workflow of
\textsc{AgRefactor}. The user provides a C/C++ program and specifies a
top-level function; from that point on, the framework is fully automated.

The pipeline begins with a \agent{Test Generator} that creates tests and interface assumptions (e.g., array sizes and function signatures) before refactoring, since the testbench must call the refactored top-level function. These assumptions propagate as high-level constraints to all downstream stages. Next, a two-stage diagnosis and planning pipeline identifies non-synthesizable constructs: category-specific \agent{Identifiers} run in parallel, an \agent{Item Organizer} merges and deduplicates their findings, and a \agent{Planner} produces a complete refactoring plan, strengthened by strategic and factual memory (Section~\ref{sec:mem}). A \agent{Refactor Worker} then applies this plan, and the result is validated through synthesis and simulation. If errors arise, an \agent{Analyzer}-\agent{Fixer} pair iteratively updates the code or testbench until both stages succeed or a retry limit is reached. Each agent is initialized with a role-specific system prompt. The \agent{Identifiers} and \agent{Planner} are further enhanced with long-term memory, so that their context is progressively enriched.

\begin{figure}
    \centering
    \includegraphics[width=0.7\linewidth]{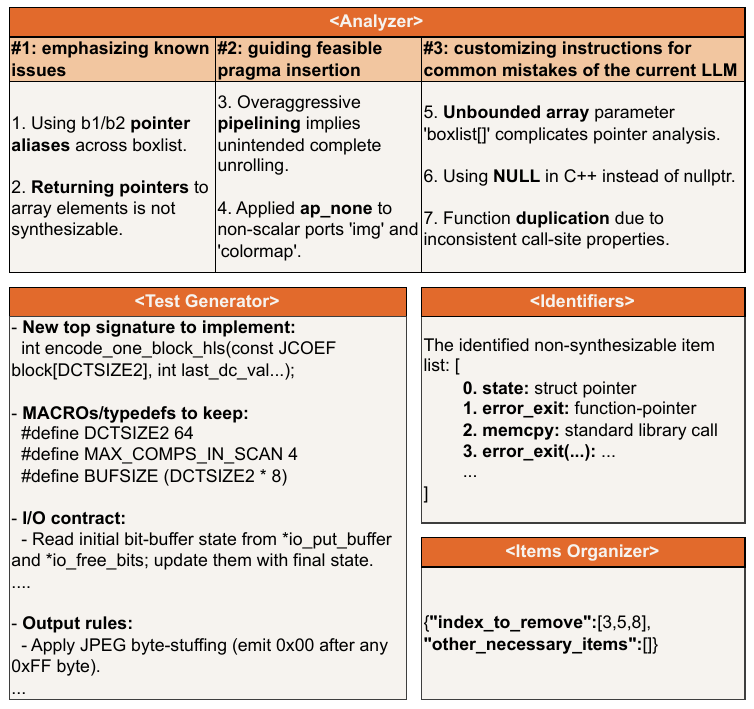}
    \caption{Messages passed between agents.}
    \label{fig:message}
\end{figure}

\subsection{Long-term Memory for HLS Refactoring}
\label{sec:mem}

To enable self-evolution, \textsc{AgRefactor} accumulates successful and unsuccessful trials as queryable knowledge\. Each memory entry is defined as a tuple $(p_{i}, I_{i}, s_{i}, c_{i})$, where $p_{i}$ is the initial program, $I_{i}$ lists the identified incompatible constructs, $s_{i}$ is the refactoring strategy, and $c_{i}$ is the generalized critique generated by the \agent{Analyzer}. During the memory accumulation phase on a diverse training set $\mathcal{P}$, the \agent{Analyzer} derives these critiques from simulation or synthesis logs. To ensure these insights are broadly applicable to unseen programs, we instruct the Analyzer to omit program-specific variables or function names. To guarantee compliance and reliability, we enforce a lightweight validation step where the Analyzer's output is cross-referenced against the original variable and function names. Critiques containing exact matches are automatically rejected and regenerated. For successful trials, we simply record the program, plan, and identified items.

When approaching a new task $q \notin \mathcal{P}$, the workflow retrieves relevant prior knowledge using a specialized distance metric that captures both global structural similarity and local implementation details. Relying solely on code embeddings compresses too much detail and may overlook non-synthesizable constructs. Thus, we combine the distance between the raw program embeddings ($e^c$) and the identified item embeddings ($e^i$):

$$d(q \notin \mathcal{P}, p \in \mathcal{P}) = \alpha|e_{q}^{c}-e_{p}^{c}|_{2}+(1-\alpha)|e_{q}^{i}-e_{p}^{i}|_{2}$$

Using this metric, we retrieve the plan $s_{i}$ from the most similar successful trials to augment the Planner's context, and the items $I_{i}$ from similar failed trials to sharpen the Identifiers' detection capabilities.

\subsection{Tool Integration}

To mitigate the high computational costs of LLMs, we integrate our framework with HeteroRefactor. Because its runtime is negligible ($<$5s), we adopt a greedy strategy: we first attempt automated refactoring with HeteroRefactor, terminating the flow immediately upon success. 

While the tool alone lacks generalizability (Section~\ref{sec:motivation}), many of its limitations are resolvable through lightweight code rewrites. If the initial attempt fails, a dedicated \agent{Tool Specialist} pipeline is triggered. The original program and identified unsupported items are passed to the \agent{Planner} and \agent{Refactor Worker} to perform tractable modifications (e.g., removing problematic headers). This targeted intervention makes the code ``HeteroRefactor-compatible,'' significantly improving overall coverage and efficiency while avoiding the overhead of full LLM-driven generation.

\subsection{Task Creation}
\label{sec:benchmark}

To construct the program set $\mathcal{P}$ for memory accumulation, we adapt existing benchmarks to Vitis HLS and extract new, standalone programs from real-world software repositories. To overcome the structural complexity of these repositories, we deploy an LLM-based agent that automates benchmark generation through four steps: selecting diverse candidate functions to maximize the coverage of non-synthesizable constructs, generating state-initializing wrappers, extracting standalone code, and validating functional correctness against the original implementation.

\subsection{Performance Optimization}

Beyond synthesizability, we develop an optimization agent that iteratively improves hardware performance. The agent operates in a CLI environment integrating C-based simulation for correctness, Vitis HLS for synthesis, and LightningSim~\cite{lightningsimv2} for fast latency estimation---bypassing time-consuming RTL simulation entirely. It also has access to multiple source-to-source code transformation tools~\cite{streamhls,prometheus}. These tools are wrapped behind a simplified command-line interface to stabilize tool calling and reduce EDA scripting effort.

The agent's system prompt enforces a strict three-phase optimization hierarchy: (1)~structural rewrites first, (2)~scheduling-report analysis to resolve critical-path violations, and (3)~fine-grained pragma tuning last. This prevents the common failure mode of naive LLM-based generators that randomly apply pragmas without first establishing a sound architecture.

To keep the agent effective over multi-hour runs, we develop a tree-based working memory pinned to its context. Each node in the tree represents a synthesized HLS code version, and the agent constructs links between nodes via condensed summaries of its optimization strategies and synthesis metrics. As the agent optimizes freely, we actively monitor both its evaluation results and context usage. To prevent context degradation, our programmatic monitors track the context window size and trigger the agent to update this tree periodically, especially before context compaction occurs. If these monitors detect performance stagnation, such as repeated simulation failures or a latency plateau, they redirect the agent to a knowledge base of successful past strategies and structural templates. This keeps the working memory compact and prevents the agent from getting stuck in
unproductive loops.
\section{Evaluation}

\label{sec:eval}

We implement \textsc{AgRefactor} using the AG2 framework~\cite{AG2_2024}. For memory storage, we rely on ChromaDB to record both successful and failed trials, and we compute embeddings for distance measurement and retrieval with the ``all-MiniLM-L6-v2'' model from the SentenceTransformer~\cite{sentence_transformers_repo} package.

\subsection{Experiment Setup}
\label{sec:eval_setup}

Our baselines for comparison are: (1) HeteroRefactor~\cite{heterorf}, the state-of-the-art automated refactoring tool, and (2) an HLSRewriter-style baseline~\cite{hlsrewriter} built on top of our own framework. Because the original HLSRewriter targets the Catapult HLS toolchain and its key components, such as the curated RAG knowledge base and identifier prompts, are not open-sourced, a direct reproduction is infeasible. Instead, we construct this baseline as a controlled ablation of AgRefactor: we replace the self-evolving memory with the official Vitis HLS programming guide as a static RAG source, collapse the six category-specific Identifiers into a single general-purpose Identifier, and remove the Tool Specialist. Critically, this baseline retains all other infrastructure: the fixing loop, synthesis/simulation pipeline, test generator, and Planner. This make it stronger than a naive reimplementation and ensuring that observed gains are attributable to our proposed components. We further ablate \textsc{AgRefactor} by comparing (i) \textsc{AgRefactor} with no memory augmentation, (ii) \textsc{AgRefactor} with only memory augmentation, and (iii) \textsc{AgRefactor} with both memory augmentation and tool integration.

The benchmark suite contains 45 cases, which we split into 24 for accumulating experience and 21 for testing the memory-augmented flow. We run the outer loop of memory accumulation for three epochs on the 24 training cases. During testing, each LLM-based workflow is executed 20 times per benchmark, and we report both the success rate and the average number of retries within the inner fixing loop. Because of the high cost of repeated evaluations, we exclude certain cases where a simple iterative flow already achieves $100\%$ success rates (Figure~\ref{fig:motivation_result}) and focus our evaluation on the more challenging benchmarks. We set $\alpha=0.6$ in all experiments.

We tested the GPT family of models. For the inner fixing flow, we cap the maximum number of iterations at three. In the memory-enabled workflow, we retrieve one plan from the most similar past task based on the distance metric defined in Section~\ref{sec:mem}, and we retrieve critiques from the top three most similar programs to enhance the context for the Identifiers. All models are run with the default parameters provided by the respective LLM APIs.

\subsection{Comparison with Baselines}

\label{sec:result_baseline}

Table~\ref{tab:baseline_comp} summarizes the results of comparing \textsc{AgRefactor} with HeteroRefactor and HLSRewriter. The improvements over HLSRewriter can be classified into three categories: (1) benchmarks that can be directly refactored by HeteroRefactor, (2) cases where the combined effort of the \agent{Identifiers}, \agent{Planners}, and memory augmentation improves accuracy on challenging benchmarks, and (3) cases where the \agent{Tool Specialist} rewrites constructs unsupported by HeteroRefactor, enabling HeteroRefactor to generate a synthesizable version through correct-by-construction transformations.

Examples of category (1) include test cases ``ahocorasick,'' ``dfs,'' and ``strassen,'' where \textsc{AgRefactor}'s tool-enhanced flow achieves the same results as HeteroRefactor with no LLM invocation. For 8 out of 11 benchmarks, our memory mechanism increases the success rate, with further details provided in Section~\ref{sec:eval_mem}. For test cases ``encode\_one\_block'' and ``idct\_generic,'' the \agent{Tool Specialist} successfully rewrites the code to match the capabilities of HeteroRefactor while maintaining functional equivalence.

On two benchmarks, ``median\_cut'' and ``argon2\_fill\_segment'', the HLSRewriter-style baseline slightly outperforms the full AgRefactor pipeline. Since both configurations share the same framework, this indicates that on certain benchmarks where a static knowledge source (the Vitis manual) already covers the relevant constructs, memory retrieval from dissimilar past tasks can introduce marginal noise.

\begin{table}[ht]
  \centering
  \caption{\small Comparison between \textsc{AgRefactor} and current SoTAs. Each benchmark is run 20 times, and we report the number of successful runs. Overall, \textsc{AgRefactor} outperforms or matches SoTA approaches on 9 out of 11 benchmarks.}
  \vspace{-5pt}
  \label{tab:baseline_comp}
  \begin{tabular}{lccc}
    \toprule
    \textbf{Task (N=20)} & \textbf{HeteroRF} & \textbf{HLSRewriter} & \textbf{Ours} \\
    \midrule
    av1\_compound\_type\_rd  & 0  & 0  & \textbf{1}  \\
    encode\_one\_block  & 0  & 2  & \textbf{10} \\
    idct\_generic   & 0  & 2  & \textbf{5}  \\
    median\_cut  & 0  & 20 & 18 \\
    argon2\_fill\_segment   & 0  & 17 & 13 \\
    mm\_chain\_dp\_orig & 0  & 5  & \textbf{6}  \\
    ahocorasick   & 20 & 16 & \textbf{20} \\
    dfs  & 20 & 16 & \textbf{20} \\
    strassen   & 20 & 17 & \textbf{20} \\
    wordbreak   & 0  & 17 & \textbf{20} \\
    skyline   & 0  & 20 & \textbf{20} \\
    \bottomrule
  \end{tabular}
\end{table}

Even for simple cases (``ahocorasick'' to ``skyline''), where a basic iterative workflow already achieves high accuracy, \textsc{AgRefactor}'s memory mechanism further improves performance, reaching nearly $100\%$. Manual inspection suggests this is because similar benchmarks appear in the memory pool, making retrieval particularly effective when the gap between ``training'' (used for memory accumulation) and ``testing'' benchmarks is small.

This also explains why improvements are limited on larger programs such as ``av1\_compound\_type\_rd''. With over 1,000 lines of code, it introduces a significant distribution shift relative to stored tasks, making past strategies and critiques less transferable.

Overall, our results highlight the value of leveraging past experience through a self-evolving agentic workflow.

\subsection{Ablation Study}

In this section, we ablate two core components of \textsc{AgRefactor}: the memory mechanism and the tool specialist. The results highlight how these enhancements extend beyond a fixed agentic workflow, providing clear advantages in both accuracy and robustness.

\subsubsection{Ablating Memory Augmentation}
\label{sec:eval_mem}

\begin{table}[h]
\centering
\small
\caption{Ablation study of \textsc{AgRefactor} with and without memory augmentation, using GPT-5-mini as the base LLM.}
\vspace{-5pt}
\begin{tabular}{l rr rr}
\toprule
 & \multicolumn{2}{c}{\textbf{no-mem}} & \multicolumn{2}{c}{\textbf{with-mem}} \\
\cmidrule(lr){2-3} \cmidrule(lr){4-5}
\textbf{Task (N=20)} & \# pass & \# retry & \# pass & \# retry \\
\midrule
av1\_compound\_type\_rd & 0 & 0.0 & \textbf{1} & 1.0 \\
encode\_one\_block & 3 & 1.0 & \textbf{5} & 1.4 \\
idct\_generic & 3 & 0.3 & \textbf{5} & 1.6 \\
median\_cut & 13 & 0.4 & \textbf{18} & 0.4 \\
argon2\_fill\_segment & 12 & 0.8 & \textbf{13} & 1.3 \\
mm\_chain\_dp\_orig & \textbf{9} & 0.3 & 6 & 0.7 \\
ahocorasick & 17 & 0.3 & \textbf{18} & 0.4 \\
dfs & 18 & 0.4 & \textbf{20} & 0.1 \\
strassen & 17 & 0.4 & \textbf{19} & 0.1 \\
wordbreak & 17 & 1.1 & \textbf{20} & 1.0 \\
skyline & 20 & 0.3 & \textbf{20} & 0.6 \\
\midrule
\emph{\% Winning (v.s. no-mem)} & \multicolumn{4}{r}{82\%} \\
\emph{\% Draw} & \multicolumn{4}{r}{0\%} \\
\bottomrule
\end{tabular}
\label{tab:result_gpt5mini_mem}
\end{table}

\begin{table}[h]
\centering
\small
\caption{Ablation study of \textsc{AgRefactor} with and without memory augmentation, using GPT-5 as the base LLM.}
\vspace{-5pt}
\begin{tabular}{l rr}
\toprule
 & \multicolumn{2}{c}{\textbf{GPT-5 (N=20)}} \\
\cmidrule(lr){2-3}
\textbf{Task} & no-mem (\# pass) & with-mem (\# pass) \\
\midrule
av1\_compound\_type\_rd & \textbf{6} & 5 \\
encode\_one\_block & 2 & \textbf{9} \\
idct\_generic & 0 & \textbf{2} \\
median\_cut & 15 & \textbf{18} \\
argon2\_fill\_segment & 14 & \textbf{18} \\
mm\_chain\_dp\_orig & 18 & 18 \\
\midrule
\emph{\% Winning (v.s. no-mem)} & \multicolumn{2}{r}{67\%} \\
\emph{\% Draw} & \multicolumn{2}{r}{10\%} \\
\bottomrule
\end{tabular}
\label{tab:result_gpt5_mem}
\end{table}

Table~\ref{tab:result_gpt5mini_mem} and Table~\ref{tab:result_gpt5_mem} report the results of \textsc{AgRefactor} with and without the proposed memory mechanism, using either the weaker ``GPT-5-mini'' or the stronger ``GPT-5'' as the base LLM. Across both models, we observe a consistent trend: adding memory augmentation improves the refactoring success rate. With GPT-5-mini, the memory-enabled workflow outperforms the no-memory baseline on 10 out of 11 benchmarks. With GPT-5, we see accuracy gains on all benchmarks overall, and the memory-enabled workflow outperforms or matches the baseline on 4 out of 6 benchmarks, showing a particularly large improvement on the ``encode\_one\_block'' benchmark.

Beyond accuracy, memory augmentation also reduces the number of retries (i.e., inner fixing iterations), such as ``ahocorasick'', ``dfs'' and ``word\_break''. This aligns with the intuition that specialized instructions stored in memory improve the robustness of the base LLM. Note that we only record the number of retry if the synthesis and simulation succeed in the end; therefore, it is possible to have larger number of retry when the success rate improves.

\begin{table*}[!t]
\centering
\caption{\small Extended HLS optimization benchmark suite (51 designs).}
\vspace{-5pt}
\small
\begin{tabular}{llr}
  \toprule
  \textbf{Source} & \textbf{Name} & \textbf{Lines of code} \\
  \midrule
  StreamHLS~\cite{streamhls} (14) & \makecell[l]{
      atax, bicg, gemm, gesummv, mvt, k2mm, k3mm, k7mm$\times$2, \\
      Autoencoder, ResidualBlock, ResMLP, DepthwiseConv, MultiHeadSelfAttention
  } & 127-990 \\ \midrule
  Prometheus~\cite{prometheus} (9) & \makecell[l]{
      atax, bicg, gemm, gemver, gesummv, mvt, symm, 2mm, 3mm
  } & 360-1050 \\ \midrule
  SoDA~\cite{soda} (6) & \makecell[l]{
      jacobi2d, jacobi3d, heat3d, seidel2d, sobel2d, denoise2d
  } & 628-1853 \\ \midrule
  HLSFactory~\cite{hlsfactory} (6) & \makecell[l]{
      3d-rendering, optical-flow, SkyNet, ECG\_Classifier, GP\_fixed, spam-filter
  } & 334-2290 \\ \midrule
  Vitis Libraries (L2)~\cite{vitislib} (16) & \makecell[l]{
      jpeg\_decoder, resize, lz4$\times$2, snappy$\times$2, geoip, kmeans, \\
      black\_scholes, swaption, portfolio, gematrixinverse, geqrf, getrf$\times$2, trtrs
  } & 140-2286 \\
  \bottomrule
\end{tabular}
\label{tab:opt_benchmark}
\end{table*}

\subsubsection{Ablating Tool Calling}
To isolate the effect of the \agent{Tool Specialist}, we conduct an ablation study comparing two settings: running HeteroRefactor alone and running HeteroRefactor with the Tool Specialist. The results, shown in Table~\ref{tab:result_tool}, demonstrate that adding the Tool Specialist, which performs simple rewrites, significantly improves success rates on ``encode\_one\_block'', ``idct\_generic'' and ``argon2\_fill\_segment''. A promising future direction is to enable LLM-based agents to directly modify or extend the capabilities of the tool itself.

\begin{table}[h]
  \centering
  \small
  \caption{Ablation study of HeteroRefactor with and without a \agent{Tool Specialist}, using GPT-5-mini as the base LLM.}
  \vspace{-5pt}
  \setlength{\tabcolsep}{6pt}
  \begin{tabular}{lrr}
    \toprule&
\multicolumn{2}{c}{\textbf{GPT-5-mini (N=20)}} \\
\cmidrule(lr){2-3}
\textbf{Task} & no-specialist & with-specialist  \\
   \midrule
    av1\_compound\_type\_r        & 0  & 0  \\
    encode\_one\_block        & 0  & 10 \\
    idct\_generic         & 0  & 5  \\
    median\_cut         & 0  & 0  \\
    argon2\_fill\_segment        & 0  & 8  \\
    mm\_chain\_dp\_orig       & 0  & 0  \\
    ahocorasick         & 20 & 20 \\
    dfs        & 20 & 20 \\
    strassen         & 20 & 20 \\
    wordbreak         & 0  & 0  \\
    skyline         & 0  & 0  \\
    \bottomrule
  \end{tabular}
  \vspace{-10pt}
  \label{tab:result_tool}
\end{table}

\subsection{Refactoring for Optimization}
The synthesizable designs produced by \textsc{AgRefactor} serve as starting points for performance optimization. We compare two approaches under a 2-hour budget: (1)~AutoDSE~\cite{autodse}, which explores pragma configurations (\textsf{PIPELINE}, \textsf{UNROLL}, \textsf{TILE}) without modifying the code structure, and (2) \textsc{AgRefactor}'s optimization agent (GPT-5.3-codex) augmented with code-transformation tools. As shown in Table~\ref{tab:performance}, the \textsc{AgRefactor} optimization agent achieves a geometric mean speedup of $6.51\times$ over AutoDSE alone. These gains stem from structural transformations inaccessible to pragma-only DSE: replacing an $O(n^2)$ sort with a radix sort (\textsf{skyline}, $295\times$), restructuring a DFS stack (\textsf{word\_break}, $29\times$), and eliminating runtime table initialization (\textsf{idct\_generic}, $10\times$). Because the agent is instructed to utilize the available pragma tuning tools initially, it matches AutoDSE's performance on benchmarks where pragma parallelism dominates (\textsf{compound\_type\_rd}, \textsf{strassen}). Furthermore, running AutoDSE for 8 hours yields performance identical to its 2-hour run, underscoring the critical importance of structural refactoring for performance optimization.

\begin{table}[h!]
\centering
\caption{\small Optimization comparison on \textsc{AgRefactor} designs, normalized to AutoDSE. Both approaches were run for 2 hours on the same refactored code. Designs that AutoDSE failed to process are excluded.}
\vspace{-5pt}
\begin{tabular}{lrr}
\toprule
\textbf{Benchmark} & \textbf{AutoDSE} & \textbf{AgRefactor} \\
\midrule
compound\_type\_rd   & $1.00\times$ & $1.00\times$ \\
strassen             & $1.00\times$ & $1.00\times$ \\
skyline              & $1.00\times$ & $295.44\times$ \\
word\_break          & $1.00\times$ & $28.71\times$ \\
idct\_generic        & $1.00\times$ & $9.95\times$ \\
encode\_one\_block   & $1.00\times$ & $6.30\times$ \\
ahocorasick          & $1.00\times$ & $4.34\times$ \\
mm\_chain\_dp        & $1.00\times$ & $1.40\times$ \\
\midrule
\textbf{Geomean}     & $1.00\times$ & $\mathbf{6.51\times}$ \\
\bottomrule
\end{tabular}
\label{tab:performance}
\end{table}

\begin{figure}
    \centering
    \includegraphics[width=0.85\linewidth]{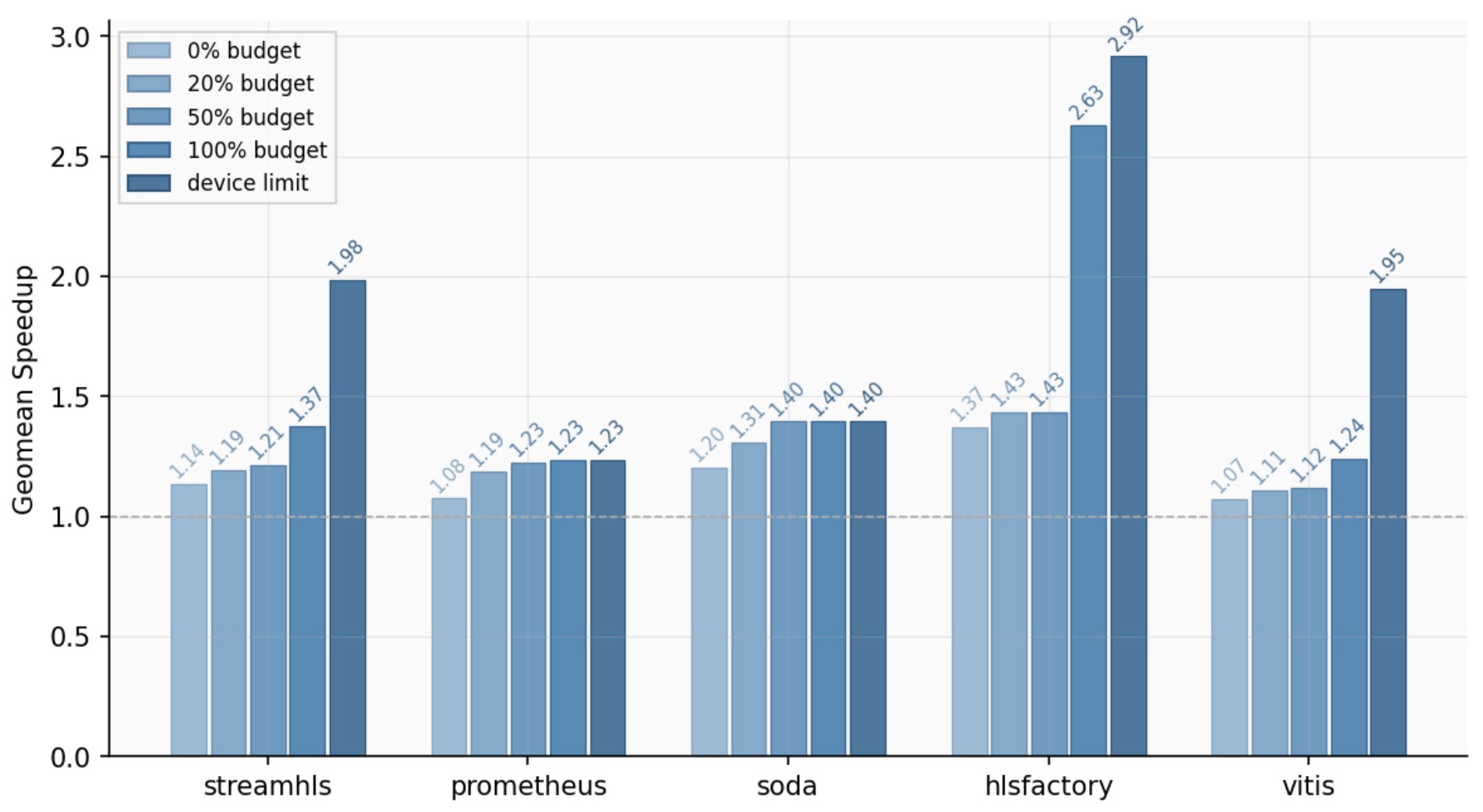}
    \vspace{-5pt}
    \caption{Performance improvement over optimized open-source designs under different extra resource budgets. Device limits are Xilinx U200 (U55C for Prometheus).}
    \label{fig:opt_grouped}
\end{figure}

We further evaluate the tool-integrated iterative optimization capability on 51 open-source HLS benchmarks spanning five categories (Table~\ref{tab:opt_benchmark}). We instruct the agent to invoke code transformation tools and then perform free-form iterative code refinement for 2 hours. During evaluation, we target the Xilinx U200 device, with the exception of the Prometheus benchmarks, which utilize the U55C setting to maintain consistency with prior work~\cite{prometheus}. Figure~\ref{fig:opt_grouped} reports the geometric mean speedup by benchmark category under increasing extra resource budgets. With no additional resource budget allowed (0\%), the agent achieves a $1.14\times$ geometric mean speedup through code restructuring alone. The techniques discovered by the agent include pure algorithmic rewrites, HLS-level register insertion, and resource binding. Relaxing the budget to 20\%, 50\%, and 100\% additional resources enables progressively more aggressive area/latency tradeoffs, reaching speedups of $1.20\times$, $1.23\times$, and $1.41\times$, respectively. The HLSFactory benchmarks benefit the most from relaxed budgets ($1.37\times \to 2.63\times$), as certain baselines are largely unoptimized and do not exploit sufficient parallelism.

\subsection{Runtime and Cost}
The runtime of our workflow depends on the LLM serving setup. (1) With commercial APIs, \textsc{AgRefactor} without tools already runs efficiently: 3-20 minutes (avg. 10), over 10x faster than manual refactoring, and only 9\% slower than HLSRewriter. (2) The tool-enabled flow further improves speed, averaging 4 minutes versus 10. (3) Tool-calling agents also reduce cost, as tool usage incurs no API token overhead.
\section{Conclusion}

We present \textsc{AgRefactor}, an agentic approach for refactoring software into HLS-compatible implementations. Our method introduces a self-evolving memory mechanism based on trial-and-error and integrates a rule-based refactoring tool to mitigate the high computational cost of LLM inference. Experimental results demonstrate superior refactoring accuracy and efficiency compared with both algorithmic and prior LLM-based methods.

\newpage
\clearpage
\nocite{openai_gpt5}
\bibliographystyle{IEEEtran}
\bibliography{main}

\newpage
\appendix

\section{Additional Ablations and Verification Analysis}
\label{sec:appendix}

\noindent\textsc{AgRefactor} is publicly available at\\
\mbox{\url{https://github.com/Williamzou0123/AgRefactor}}

This appendix presents supplementary studies that complement the main evaluation: the effect of self-accumulating memory across training epochs, the contribution of the LLM-based tool specialist on the rule-based tool path, and a held-out analysis of testbench strength. Unless otherwise stated, all experiments use GPT-5-mini as the base model and pool $N=20$ attempts per kernel.

\subsection{Self-Accumulating Memory Across Epochs}

\textsc{AgRefactor}'s memory is \emph{persistent across tasks}: knowledge distilled while refactoring one kernel is stored and later retrieved when refactoring subsequent, previously unseen kernels. To isolate the effect of this accumulation, Table~\ref{tab:ablation_memory_epochs} compares three configurations under an identical pipeline: no memory (\textbf{no-mem}), memory after a single epoch of accumulation (\textbf{epoch=1}), and memory after three epochs (\textbf{epoch=3}). The total number of successful refactors rises monotonically from 129 to 133 to 145, a 12.4\% relative improvement over the memoryless baseline, and the gains are distributed across most kernels rather than concentrated in a few. A small number of kernels (e.g., \texttt{mm\_chain\_dp\_orig}) show no benefit, reflecting run-to-run variance and limited cross-task transfer for the most dissimilar program. To our knowledge, cross-task persistent memory of this form has not previously been studied for refactoring agents; existing memory mechanisms for coding agents largely operate within a single task.

\subsection{Contribution of the LLM-based Tool Specialist}

\textsc{AgRefactor} does not invoke the rule-based refactoring engine (HeteroRefactor) naively. Before the tool is called, an LLM-based \emph{tool specialist} performs a preliminary rewrite that maps constructs unsupported by the rule-based engine onto a compatible form. Table~\ref{tab:ablation_heterorefactor} isolates this contribution by comparing the naive LLM flow against the same flow extended with the preprocessor and the tool path. The tool path increases the total number of successful refactors from 129 to 146 (+17, a 13.2\% relative gain), with the largest improvements on \texttt{encode\_one\_block} (3 to 10) and on \texttt{ahocorasick}, \texttt{dfs}, and \texttt{strassen} (each reaching 20). The deterministic tool and the LLM are thus complementary: the specialist makes substantially more programs amenable to fast, reliable rule-based transformation, while the LLM handles cases the tool cannot.

\subsection{Testbench Strength and Held-out Re-evaluation}

Because \textsc{AgRefactor}'s correctness signal is provided by LLM-generated testbenches, we analyze their strength directly. In the original flow each testbench is generated in a single shot. To quantify how reliable this signal is, we perform a held-out re-evaluation on the four kernels whose single-shot testbenches achieved the lowest coverage of the original source. For every refactor accepted by the original (public) testbench, we re-run it against a stronger, coverage-optimized (hidden) testbench that preserves the same interface signature. Table~\ref{tab:inflation_gap} reports the outcome: of the attempts accepted by the public testbench (15 without memory and 17 with memory), only 6 and 7 respectively also pass the stronger testbench, an inflation gap of roughly 60\% in both settings. Single-shot testbenches can therefore be lenient on the hardest kernels, which motivates a stronger evaluator.

To close this gap we augment the testbench generator with an engineer--rater loop. A rater agent audits each candidate testbench before use: it instruments the testbench to log the reference function's outputs and confirms that they are non-trivial and vary across inputs, and it synthesizes a deliberately incorrect (``cheating'') refactor to verify that the testbench rejects it. The loop explores several trajectories per kernel and retains the testbench with the strongest coverage. On the four weakest kernels this raises coverage of the original source on \texttt{av1\_compound\_type\_rd} (60.1\% to 87.3\%), \texttt{encode\_one\_block} (79.6\% to 100.0\%), \texttt{idct\_generic} (46.9\% to 100.0\%), and \texttt{mm\_chain\_dp\_orig} (45.5\% to 89.9\%). The remaining seven kernels were already well exercised by the single-shot generator (90.0--100.0\% coverage) and were left unchanged.

Following much of the RTL/HLS generation literature, we treat a refactor as successful when it passes its testbench. Stronger functional verification is an important and largely orthogonal direction; the central contribution of this work is the self-accumulating, cross-task memory for refactoring agents, with the rule-based tool path and tool specialist providing complementary, low-cost reliability gains.

\begin{strip}
\centering
\small
\captionof{table}{Ablation study of \textsc{AgRefactor} with and without memory augmentation, using GPT-5-mini as the base LLM. All columns pool $N=20$ per kernel.}
\vspace{-5pt}
\begin{tabular}{l rr rr rr}
\toprule
 & \multicolumn{2}{c}{\textbf{no-mem}} & \multicolumn{2}{c}{\textbf{with-mem (epoch=1)}} & \multicolumn{2}{c}{\textbf{with-mem (epoch=3)}} \\
\cmidrule(lr){2-3} \cmidrule(lr){4-5} \cmidrule(lr){6-7}
\textbf{Task (N=20)} & \# pass & \# retry & \# pass & \# retry & \# pass & \# retry \\
\midrule
av1\_compound\_type\_rd      &      0 &  0.0 &      0 &  0.0 & \textbf{1} &  1.0 \\
encode\_one\_block           &      3 &  1.0 &      3 &  1.7 & \textbf{5} &  1.4 \\
idct\_generic                &      3 &  0.3 &      4 &  0.5 & \textbf{5} &  1.6 \\
median\_cut                  &     13 &  0.4 & \textbf{18} &  0.6 & \textbf{18} &  0.4 \\
argon2\_fill\_segment        &     12 &  0.8 &     12 &  1.0 & \textbf{13} &  1.3 \\
mm\_chain\_dp\_orig          & \textbf{9} &  0.3 &      8 &  0.1 &      6 &  0.7 \\
ahocorasick                  &     17 &  0.4 &     17 &  0.2 & \textbf{18} &  0.4 \\
dfs                          &     18 &  0.3 & \textbf{20} &  0.2 & \textbf{20} &  0.1 \\
strassen                     &     17 &  0.2 &     16 &  0.0 & \textbf{19} &  0.1 \\
wordbreak                    &     17 &  1.1 &     16 &  0.6 & \textbf{20} &  1.0 \\
skyline                      & \textbf{20} &  0.2 &     19 &  0.2 & \textbf{20} &  0.6 \\
\midrule
\emph{Total pass} & \multicolumn{1}{r}{129} & & \multicolumn{1}{r}{133} & & \multicolumn{1}{r}{145} & \\
\bottomrule
\end{tabular}
\label{tab:ablation_memory_epochs}
\end{strip}


\begin{table*}[t]
\centering
\small
\caption{Effect of adding the preprocessor + heterorefactor tool path on top of the naive LLM flow, using GPT-5-mini. Column 2 reports the per-kernel union with the tool path. All columns pool $N=20$ per kernel.}
\vspace{-5pt}
\begin{tabular}{l rr rr}
\toprule
 & \multicolumn{2}{c}{\textbf{naive LLM}} & \multicolumn{2}{c}{\textbf{naive LLM + preproc+heteroRF}} \\
\cmidrule(lr){2-3} \cmidrule(lr){4-5}
\textbf{Task (N=20)} & \# pass & \# retry & \# pass & \# retry \\
\midrule
av1\_compound\_type\_rd      &      0 &  0.0 &      0 &  0.0 \\
encode\_one\_block           &      3 &  1.0 & \textbf{10} &  0.0 \\
idct\_generic                &      3 &  0.3 & \textbf{5} &  0.0 \\
median\_cut                  & \textbf{13} &  0.4 & \textbf{13} &  0.4 \\
argon2\_fill\_segment        & \textbf{12} &  0.8 & \textbf{12} &  0.8 \\
mm\_chain\_dp\_orig          & \textbf{9} &  0.3 & \textbf{9} &  0.3 \\
ahocorasick                  &     17 &  0.4 & \textbf{20} &  0.0 \\
dfs                          &     18 &  0.3 & \textbf{20} &  0.0 \\
strassen                     &     17 &  0.2 & \textbf{20} &  0.0 \\
wordbreak                    & \textbf{17} &  1.1 & \textbf{17} &  1.1 \\
skyline                      & \textbf{20} &  0.2 & \textbf{20} &  0.2 \\
\midrule
\emph{Total pass} & \multicolumn{1}{r}{129} & & \multicolumn{1}{r}{146} & \\
\emph{Delta vs naive LLM} & \multicolumn{2}{r}{--} & \multicolumn{2}{r}{\textbf{+17 (+13.2\%)}} \\
\bottomrule
\end{tabular}
\label{tab:ablation_heterorefactor}

\vspace{16pt}

\caption{Held-out testbench re-evaluation on the four lowest-coverage kernels. For each refactoring attempt accepted by the original single-shot testbench, we coverage-optimize that testbench (preserving its interface signature) and re-run the refactored code against the resulting stronger testbench. ``Inflated\textsubscript{mismatch}'' counts attempts that the original testbench marked as passing but for which the stronger testbench exposes an output mismatch. All conditions pool $N=20$ attempts per kernel across two independent runs.}
\vspace{-5pt}
\begin{tabular}{l rrr rrr}
\toprule
 & \multicolumn{3}{c}{\textbf{no-mem}} & \multicolumn{3}{c}{\textbf{with-mem}} \\
\cmidrule(lr){2-4} \cmidrule(lr){5-7}
\textbf{Task} & \# pass\textsubscript{public} & \# pass\textsubscript{hidden} & \# inflated\textsubscript{mismatch} & \# pass\textsubscript{public} & \# pass\textsubscript{hidden} & \# inflated\textsubscript{mismatch} \\
\midrule
av1\_compound\_type\_rd      &   0 &   0 &   0 &   1 &   0 &   1  \\ 
encode\_one\_block           &   3 &   0 &   2 &   5 &   2 &   4  \\ 
idct\_generic                &   3 &   0 &   3 &   5 &   4 &   3  \\ 
mm\_chain\_dp\_orig          &   9 &   6 &   3 &   6 &   1 &   4  \\ 
\midrule
Total                        & \textbf{15} & \textbf{6} & \textbf{8} & \textbf{17} & \textbf{7} & \textbf{12} \\
Inflation gap                & \multicolumn{2}{r}{9 (60\%)} &  & \multicolumn{2}{r}{10 (59\%)} &  \\
\bottomrule
\end{tabular}
\label{tab:inflation_gap}
\end{table*}

\end{document}